\documentclass[10ptm,margin=1in]{article}
\usepackage[utf8]{inputenc}
\usepackage[ margin=1in]{geometry}
\usepackage[T1]{fontenc}
\usepackage{hyperref}
\usepackage{graphicx} 
\usepackage{epsfig} 
\usepackage{amsmath} 
\usepackage{amssymb}
\usepackage{soul}
\usepackage{url}
\usepackage{xcolor}
\usepackage{subcaption}

\usepackage{float}
\usepackage{authblk}
\usepackage[linesnumbered,ruled,vlined]{algorithm2e}
\usepackage{mathtools}

\date{\vspace{-5ex}}

\usepackage[
    backend=biber,
    style=numeric,
    dateabbrev=false,
    url=true, 
    doi=true,
    eprint=false,
    sorting=nty,
    giveninits=true,
    urldate=iso,
    maxbibnames=50
]{biblatex}
\SetCommentSty{mycommfont}

\title{Collaborative Robot Arm Inserting Nasopharyngeal Swabs with Admittance Control}

\author[1,*]{Peter Q. Lee}
\author[1]{John S. Zelek}
\author[1,2,3]{Katja Mombaur}
\affil[1]{Systems Design Engineering, University of Waterloo, 200 University Avenue West, Waterloo, Canada}
\affil[2]{Mechanical and Mechatronics Engineering, University of Waterloo, 200 University Avenue
  West, Waterloo, Canada}
\affil[3]{Optimization and Biomechanics for Human-Centred Robotics (BioRobotics Lab), Institute for Anthropomatics and Robotics, Karlsruhe Institute of Technology, Karlsruhe, Germany}
\affil[*]{pqjlee@uwaterloo.ca}

\begin{document}
\flushbottom
\maketitle

\begin{abstract}
The nasopharyngeal (NP) swab sample test, commonly used to detect COVID-19 and other respiratory
illnesses, involves moving a swab through the nasal cavity to collect samples from the nasopharynx. While
typically this is done by human healthcare workers, there is a significant societal interest to enable robots to do this test to reduce exposure to patients and to free up
human resources. The task is challenging from the robotics perspective because of the dexterity and
safety requirements. While other works have implemented specific hardware solutions, our research
differentiates itself by using a ubiquitous rigid robotic arm. This work presents a case study where we investigate the strengths and
  challenges using compliant control system to accomplish NP swab tests with such a robotic configuration.
  To accomplish this, we designed a force sensing end-effector that integrates with
  the proposed torque controlled compliant control loop. 
  We then conducted experiments where the robot inserted NP swabs into a 3D printed nasal cavity
  phantom. Ultimately, we found that the compliant control system outperformed a basic position
  controller and shows promise for human use. However, further efforts are needed to ensure the
  initial alignment with the nostril and to address head motion.
\end{abstract}

\section{Introduction}
\label{sec:intro}
From the outset of the COVID-19 pandemic, the appeal to applying robots in the place of human
healthcare workers has increased substantially. A major occupational hazard for healthcare workers
is contracting illnesses from the patients they are treating; especially via highly contagious
air-born spread illnesses like COVID-19. Robots deployed in healthcare have the 
advantage of being immune to illnesses, and would thereby be useful to protecting the healthcare workers
and preventing downtime~\cite{Kaiser2021-312}. In cases where training consistency is a concern~\cite{Liu2021-317}, robotics can provide a way to standardize care. However, close-contact healthcare tasks are challenging workspaces for robots because of safety considerations and the need to meet medical objectives
with the constraints of the robotic hardware.

Consequently, in this research we target the task of nasopharyngeal (NP) swab sample collection
using robots.  The task involves inserting a thin, flexible
swab through the nasal cavity until it reaches the nasopharynx at the
anterior (just above the back of the throat). The task is challenging
because the swab has to navigate around anatomical obstacles in the 
nasal cavity; namely the nasal septum (the wall that divides the left and right passages of the
nose) and the inferior turbinates (the curved bony structures). The ideal path for the swab to
travel is between the nasal palate and the inferior turbinates~\cite{Piras2020-169} (example also later
  shown in Fig. \ref{fig:apparatus}); deviation from this path is unlikely
to reach nasopharynx and can cause discomfort or injury if the swab impacts sensitive anatomy such
as cribriform plate and the various nerve clusters in the nasal cavity~\cite{Sobiesk2021-186}. From a control perspective, it is an interesting task because once the
swab enters the nose, it becomes visually unobservable, so any adjustments must be based on measured forces
applied to the swab.

\begin{figure}\centering
  \includegraphics[width=0.75\linewidth]{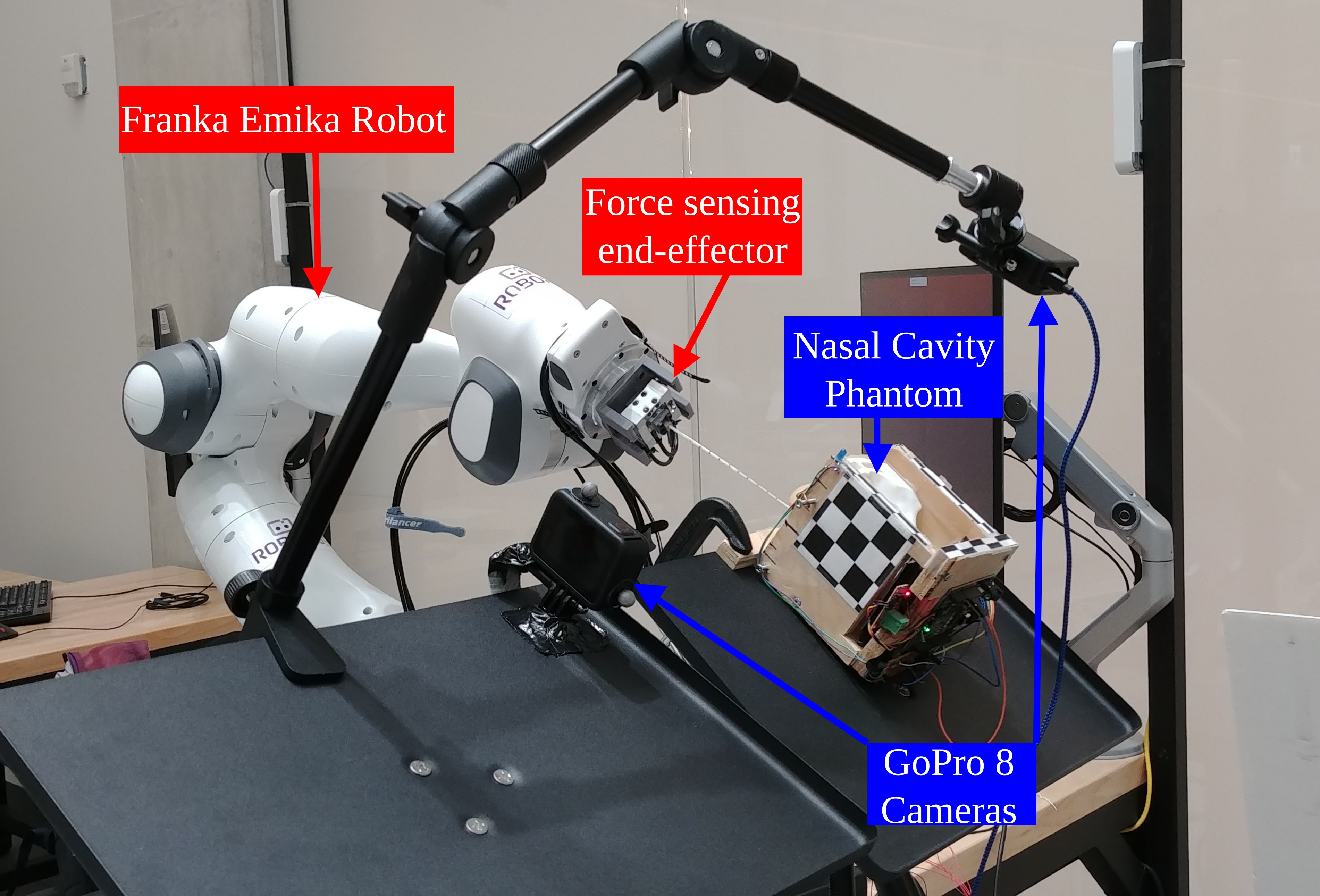}
  \caption{Nasal cavity apparatus arranged next to the Franka Emika Robot arm with the proposed force sensing
    swab end-effector attached. Two GoPro cameras are used to observe the outcome of the insertions
    and are not part of the control loop.}
  \label{fig:setup}
\end{figure}

Applying robots to different types of swab sampling tasks has received increasing attention from the
robotics and biomedical engineering community. Park \textit{et al.}~\cite{Park2021-301} conducted a study where the forces of a
practitioner were recorded using a handheld instrument on a phantom. Hwang \textit{et al.}~\cite{Hwang2022-310} implemented a visual-servo and control system, which uses deep-learning models
to detect the nose and guide the robot to more accurate positioning at the nostril prior to
insertion. Li \textit{et al.}~\cite{Li2023-343} also implemented visual-servo methods for NP swabs using a hierarchical decision network strategy. Li \textit{et al.}~\cite{Li2021-311} designed a 3-DOF robot with an endoscopic camera to perform oropharyngeal swab
tests via teleoperation. Chen \textit{et al.}~\cite{Chen2022-313} designed another teleoperated soft-rigid hybrid robot that
features a compliant fiber Bragg grating integrated manipulator to adapt to disturbances during
nasopharyngeal or oropharyngeal swab tests. Maeng \textit{et al.}~\cite{Maeng2022-314} created a custom NP sampling robot 
featuring a remote center of motion mechanism in order to compensate for sudden forces by the
patient during the procedure. Zhang \textit{et al.}~\cite{Zhang2023-355} created a platform centred around a humanoid dual-arm
 robot for NP swab sampling. Their system utilized a combination of RGB-D cameras, LIDAR scanners, and a force-torque sensor to
perform the task on humans, with results indicating it was very effective at gathering PCR samples.
There are also some commercial attempts at robotic 
  swabbing. Lifeline robotics~(Odense, Denmark) take on the related, but distinct task of oropharyngeal
  swabbing using a UR5 robot\cite{2021-345}. Franka Emika~(Munich, Germany) built a swabbing station surrounding their robot, but was applied towards shallow nose and throat
  swabs\cite{America2021-344}. Brain Navi~(Zhubeai, Taiwan) applied a UR5 robot to do NP swabs based on a
  3D visual scan of the face, but appears to do the test without any force sensing feedback~\cite{2020-346}.

Many of the developments listed above have integrated or developed custom robotic hardware in order
to do the swabbing test. 
Instead, we consider an alternative scenario where a single collaborative arm could be delegated
  to a multitude of healthcare tasks that require close contact with a patient (e.g., temperature
  testing, needlework, sample collection). Specifically, we look to use a rigid manipulator robot to
  autonomously collect NP swab samples, under the scenario that the hardware could generalize to
  other tasks by utilizing the same instruments that a human worker would use. While the full NP test
  consists of three stages, a) inserting the swab through the nasal cavity until it reaches the nasopharynx b)
  rotating the swab on the nasopharynx c) removing the swab from the nasal cavity~\cite{Piras2020-169}, this research
  focuses on achieving the first stage of the test. The constraints from this scenario provide an
  interesting challenge because the contact forces in these tasks would need to be regulated by the
  control law rather than relying on mechanical compliance from specifically designed
  manipulators. With respect to these challenges, we contribute by conducting a case study to examine
    the suitability of a robotic arm in this scenario using an admittance control system. We first
    describe our design of a low-cost, high accuracy force sensing end-effector that fits onto the
  flange of a collaborative robotic arm. We then describe our design of a system featuring a
  force-feedback torque control law to execute the insertion stage of the NP swab test. Finally, we engage in
  experiments with the robot on a 3D printed phantom of a human nasal cavity and compare its
  performance with a baseline position controller to determine the effectiveness of the setup for
  performing NP swab tests.

\section{Materials and Methods}
\label{sec:hardware}
The hardware used in this study consist of the collaborative robotic arm platform, the custom designed force-sensing
end-effector where the NP swab is mounted, and the nasal cavity apparatus.

\subsection{Collaborative Robot Arm}
\label{subsec:pandaarm}
The robotic platform used in this work is the Franka Emika Robot ``Panda'' arm (Franka Emika, Munich
Germany), which we assume to follow the dynamics~\cite{Featherstone2008-217}
\begin{equation}
  \tau = M(q) \ddot{q} + C(q,\dot{q}) \dot{q} + g(q),
  \label{eq:robot_dynamics}
\end{equation}
where $M$ is the mass inertia matrix, $C$ is the Coriolis effects matrix, $g$ is the gravitational vector, $\tau$ is the torque vector, $\ddot{q},\dot{q},q$ are the joint acceleration, velocity, and position vectors. The arm has 7-DOF and is designed as a collaborative robot that is meant to be used to perform tasks in conjunction with humans. 
We implement our control method, described in \autoref{subsec:torque_control}, using the franka-ROS
API
with the torque controller interface, which provides all of the model parameters listed above. While we chose to use the Panda due to its ubiquity,
it should be noted that our work could generalize to similar collaborative arms on the market.

\subsection{Force sensing end-effector}
\label{subsec:endeffector}
While the Panda has torque sensors on each of its joints, it quickly became clear that these were
ill-suited to measure the relatively small forces applied to the NP swab. 
We moved the arm without an end-effector along the trajectory described in Section \ref{subsec:waypoint} and
  observed that the measured torques projected as force onto the end-effector, shown in
  Fig. \ref{fig:FR_noise}, reach noise levels between 40 mN to 60 mN standard deviations. In addition, model errors cause non-stationary drift because when
  the arm changes configuration the uncompensated weight of the links get erroneously interpreted as
  external torque. Under these conditions, it is clear that any low magnitude forces that would be
  transmitted through a swab would be swallowed up by these disturbances if we were to solely rely
  on the built-in torque sensors. Therefore, we designed a custom end-effector to mount a NP swab and sense the 3 axial
forces applied to it. We integrate a GPB160 10 N capacity tri-axial strain gauge loadcell (Galoce, Shaanxi, China)
onto a 3D printed housing to be mounted onto the Panda's flange. A 3D printed mount that fits
the end of an NP swab is affixed
to the exterior facing side of the loadcell (The 3D print STL files are available in the
  supplementary data). We adopt a fairly simple electronics setup: each of the
leads from the load cell axes are soldered into an HX711 Wheatstone bridge amplifier set to 80 Hz mode that is interfaced with an
Arduino Nano. An image of these components is shown in Fig. \ref{fig:end_effector}. A ROS node was
created to publish the values sensed from the three axes via the controlling computer. Overall, the
entire cost of the materials for the sensorized end-effector was less than \$300 USD, which makes it an
affordable solution, provided a robot arm is already available. 
The base of the end-effector was 3D printed with PLA to screw into the Panda's flange (DIN ISO 9409-1-A50). This base piece could easily be adapted and reprinted to other mounting flanges on other robots. While this prototype is made with swabs as the application, we foresee that other tools could be attached to the loadcell that could enable other medical tasks.

In practice, the implementation is sufficiently sensitive because of the fine capacity of the
loadcell. The electronic noise present in the signal remains quite low as well, having a standard
deviation of about 1 mN that is further low pass filtered for reasons described in
  \autoref{subsec:force_filter}. However, one detail that requires special attention is the impact
of 
gravity on the readings. As the end-effector changes orientation, the forces due to gravity will
need to be subtracted within the end-effector frame in order to isolate the raw forces applied to
the swab. In an ideal case, we would subtract the weight attached to the loadcell. However,
the  wires leading from the loadcell create non-trivial effects on the readings as
their tension changes with gravity.

The net force read on the loadcell can be described as
\begin{equation}
  F_{net} = F + G(o) + Z
\end{equation}
where $F$ is the external force vector applied to the swab, $G$ is the force gravity vector as a
function of the end-effector orientation $o=(o_x,o_y,o_z)$, and $Z$ is the bias offset. We
devised a calibration routine where the end-effector is moved to nine different orientations
with no external force ($F=0$). 
The gravitational force is estimated with a multiple linear regression model. The regression
analysis found significant effects by the non-zero terms in the $A\in \mathbb{R}^{3 \times 4}$
matrix written below, with one significant interaction term $o_y o_z$ for $G_y$.
\begin{equation}
  G(o) = \left[\begin{smallmatrix} A_{x,x} & A_{x,y} & A_{x,z} & 0  \\
    A_{y,x} & A_{y,x} & A_{y,z} & A_{y,yz}  \\
    0 & 0 & A_{z,z} & 0 \end{smallmatrix}\right] \left[\begin{smallmatrix} o_x\\ o_y\\
    o_z \\ o_{y}o_z \end{smallmatrix}\right] = A \left[\begin{smallmatrix} o_x\\ o_y\\
    o_z \\ o_{y}o_z \end{smallmatrix}\right].
\end{equation}
Finally, we can solve for the parameters of the gravitational + offset model via a least squares
problem over the $F_{net}$ and $o$ values from the nine orientations
\begin{equation}
  \min_{A,Z} {|| F_{net} - (A\begin{bmatrix} o_x& o_y&
    o_z & o_{y}o_z \end{bmatrix}^{\top} + Z) ||}^2
\end{equation}
The fit for this calibration
procedure is quite good, achieving $R^2 > 0.9999$, indicating that this method can effectively
eliminate the gravitational effects.

\begin{minipage} {0.47\linewidth}\centering
  \includegraphics[width=\linewidth]{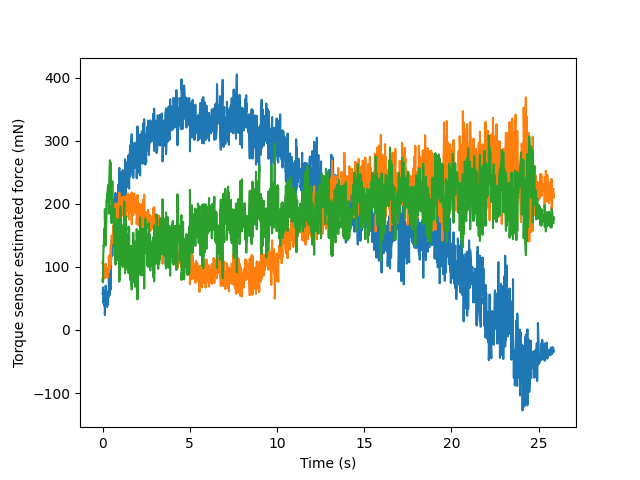}
  \captionof{figure}{Observed noisy task-space force measurements when using the FR internal torque
    sensors. This demonstrates the necessity of designing an end-effector with an external loadcell
    that is sensitive enough to observe forces transmitted by the swab.}
  \label{fig:FR_noise}
\end{minipage}\hspace{0.5cm}\begin{minipage}{0.47\linewidth}\centering
  \includegraphics[width=0.85\linewidth]{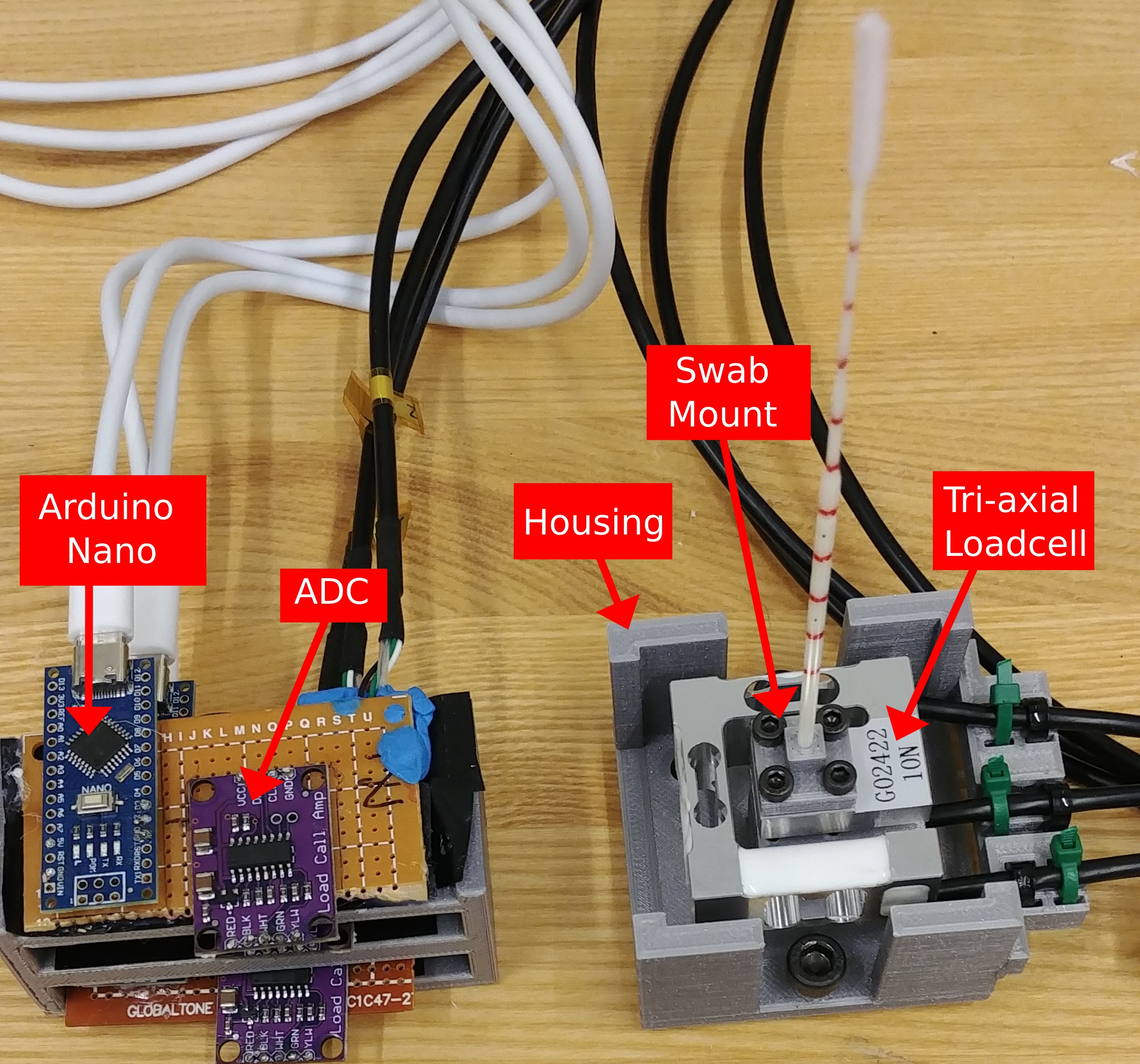}
  \captionof{figure}{Components of the custom end-effector. Left: Wheatstone bridge amplifier and digital
    conversion circuit. Right: the tri-axial loadcell interfaced with its
    housing and swab mount. Note that the red stripes on the swab are labelled to gauge distance
  in experiments, but do not have any functional purpose related to the controller.}
  \label{fig:end_effector}
\end{minipage}

\subsection{Nasal cavity apparatus and experiment setup}
\label{subsec:apparatus}
\begin{figure}
  \includegraphics[width=\linewidth]{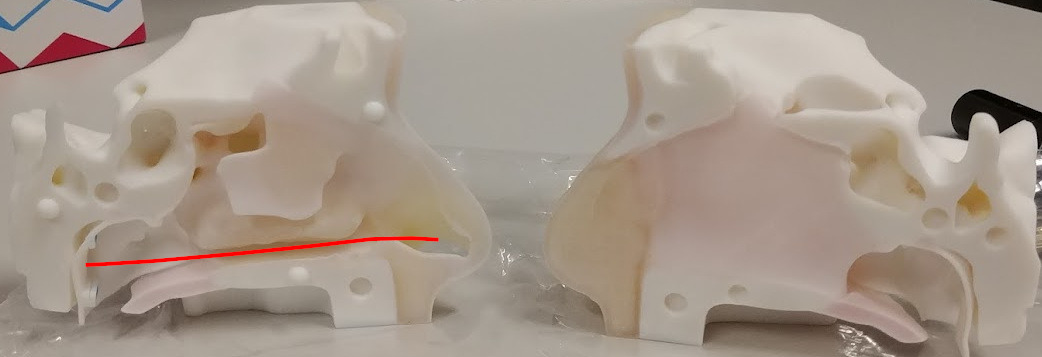}
  \caption{Nasal cavity phantom used for validation experiments. The red line highlights the path to reach the nasopharynx from the nostril.}
  \label{fig:apparatus}
\end{figure}

The other component used in this work is the nasal cavity phantom that is used to evaluate the proposed
methods. We use the 3D printed nasal cavity phantom (shown in Fig. \ref{fig:apparatus}) designed by
Sanan{\`e}s \textit{et al.}~\cite{Sananes2020-91}, that was printed with a 
PolyJet 3D printer using the rubber-like Agilus 30 to print the fleshy tissue within the nasal
cavity. The figure also shows the ideal path a swab should follow, leading from the nostril to the
nasopharynx. The phantom is placed into an open fitting 
container and was augmented by gluing a
force sensing resistor (FSR) onto the nasopharynx.
The apparatus was clamped to a tripod so that it could be re-positioned and reoriented freely. Two
GoPro Hero 8 cameras were used to record the experiments and were aligned with an adjacent tripod such that they faced the apparatus in
the top-down and sideways directions. Fig. \ref{fig:setup} shows a photo of this setup.

\subsection{Control system }
\label{sec:methods}
\begin{figure}
  \includegraphics[width=\linewidth]{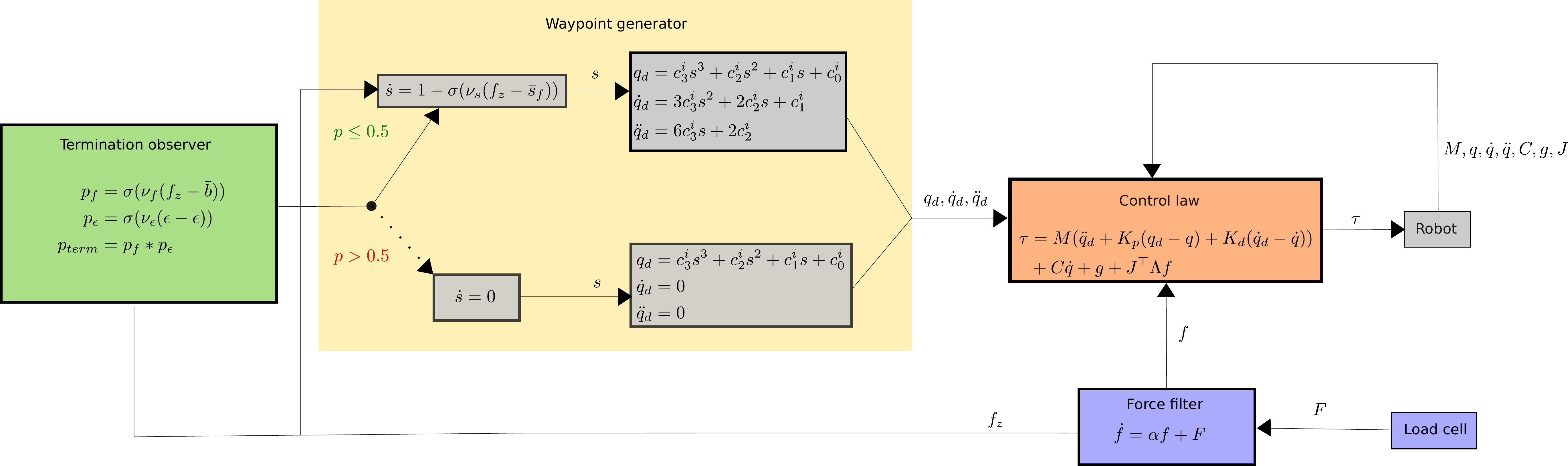}
  \caption{Block diagram for the proposed NP swab insertion system.}
  \label{fig:sys_diagram}
\end{figure}

The proposed system consists of four components: a force filter for the loadcell readings, a waypoint
trajectory generator, the torque control law, and an observer to determine when the insertion
procedure is completed. A block diagram summarizing the interaction between these components is
shown in Fig. \ref{fig:sys_diagram}. Each of these components are described in the following sections.

\subsection{Force filtering}
\label{subsec:force_filter}
\begin{figure}\centering
  \includegraphics[width=0.5\linewidth]{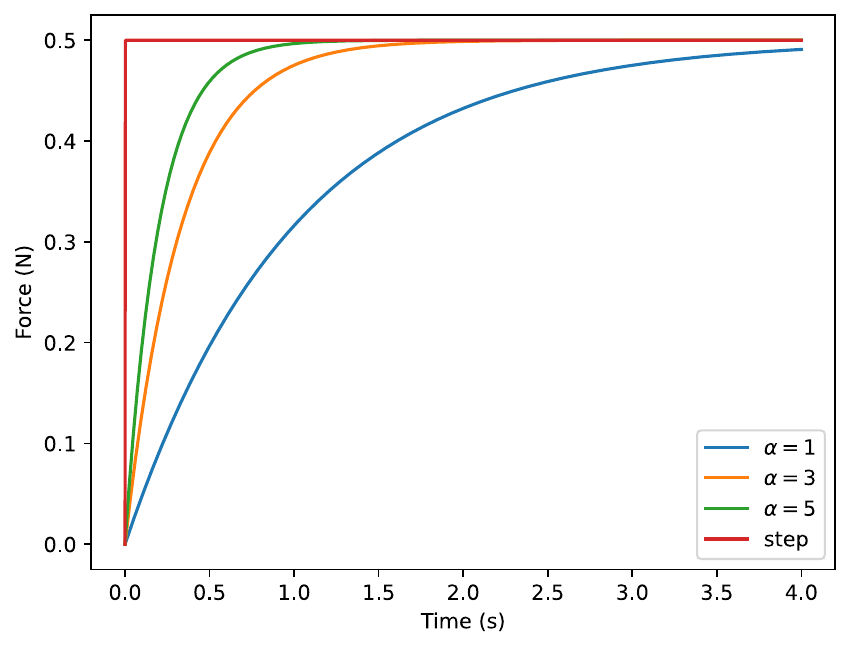}
  \caption{Example of the force filter response to a 0.5 N step function with $\alpha=1$, $\alpha=3$, and $\alpha=5$.}
  \label{fig:step_response}
\end{figure}

As described in \autoref{subsec:endeffector} the loadcell force readings already have a low
level of noise. However, it is still desirable to filter the force signals because
of the impulses that occur from intermittent contacts between the swab and the nasal cavity. If a
force-feedback controller were to react directly to the impulses, motion becomes jerky and
unstable as it rebounds between these intermittent contacts, which is undesirable and dangerous
behaviour to have within the nasal cavity. 
We therefore use the filter
\begin{equation}
  \dot{f} =  -\alpha f + \alpha F,
\end{equation}
where $F$ is the raw force measurement coming from the loadcell, $f$ is the filtered signal that
will be used in the control system, and $\alpha \ge 0$ is the response rate. This is a continuous
implementation of an exponential moving average filter; Fig. \ref{fig:step_response} shows an example
of how $\alpha$ influences the step response. Ultimately, we want the filter to attenuate sudden
changes in force and respond to consistent levels of force. Having a delayed response in $f$ is
ideal because this encourages the controller to make gradual adjustments and discourage jerky and
unstable motion. In our experiments we chose $\alpha=1$ because it provided the response we
desired.

\subsection{Waypoint generation}
\label{subsec:waypoint}
From our previous work~\cite{Lee2022-309}, we studied the deformation of NP swabs as they were inserted through the nasal
cavity and found the optimal linear trajectory to insert at to minimize swab deformation. We
build on these ideas by projecting a linear task-space trajectory to follow in the arm's
workspace. We take insertion at a 28 degree decline angle (c.f. \cite{Marty2020-170}) over 20 cm, selecting 32 points along this line
and solving for continuous joint configurations ($q$) with inverse-kinematics using
RBDL-Casadi~\cite{Michaud2023-316}. Cubic splines were fit between each of the points, having coefficients
$\{c^i\}$ with respect to the path parameter $s$. The desired nominal joint waypoints are computed
as
\begin{equation}
  \begin{aligned}
  q_d &= c^i_3 s^3 + c^i_2 s^2 + c^i_1 s + c^i_0\\
  \dot{q}_d &= 3 c^i_3 s^2 + 2 c^i_2 s + c^i_1\\
  \ddot{q}_d &= 6c^i_3 s + 2c^i_2, \\
  \end{aligned}
\end{equation}
where $c^i$ is the coefficient vector that corresponds to the spline within the domain of
$s$. These splines are all computed offline, and the coefficients are used during runtime to
  determine the associated goal positions, velocities, and accelerations.

We design the progression of $s$ with respect to time as
\begin{equation}
  \dot{s} = 1 - \sigma ( \nu_s(f_z - \bar{s}_f)),
  \label{eq:s}
\end{equation}
where $\sigma$ is the sigmoid function; the scalars $\nu_s$ and $\bar{s}_z$ influence the scale and
offset of the sigmoid response. The purpose of this formulation is to allow the nominal trajectory
to respond to disturbances via the force reading $f_z$, which faces the direction the swab will be moving along. When there is little $f_z$, the sigmoid
remains unsaturated and the trajectory proceeds normally. When there is high $f_z$, the sigmoid
becomes saturated and the trajectory slows down to allow time for the controller, described in the
\autoref{subsec:torque_control}, to make adjustments and respond to the disturbance. We set
$\bar{s}_f=0.33$ and $\nu_s = 12$; based on experimentation, these values allowed the trajectory to
slow down on when it encountered early contacts without stopping completely.

\subsection{Torque Control law}
\label{subsec:torque_control}
The task of the control law is to strike a balance between two things: a) following the nominal
trajectory waypoints, $q_d,\dot{q}_d,\ddot{q}_d$ and b) adjusting to contact forces $f$ that are
applied to the swab. As a result, we designed the admittance controller based on a
computed torque control law \cite{An1987-315} using feedforward acceleration with position, velocity, and
force feedback
\begin{equation}
    \tau = M(q) ( \ddot{q}_d + K_p (q_d - q) + K_d (\dot{q}_d - \dot{q})) + C(q,\dot{q})\dot{q} + g(q) +   J^{\top} \Lambda f.
  \label{eq:torque_control}
\end{equation}
Here, $\ddot{q}_d,\dot{q}_d,q_d$  and $\dot{q},q$ are the acceleration, velocity, and
positions for the nominal joint waypoints and the actual joints, respectively. $M$ is the mass
inertia matrix, $C$ is the Coriolis matrix, $g$ is the gravitational vector, $J$ is the $7\times 3$ joint-euclidean space Jacobian matrix,
$ K_p=\text{diag}(600,600,600,600,600,600,50),K_d=\text{diag}(30,30,30,30,30,30,5)$ are diagonal
gain matrices for position and velocity.
One can see how the first line is used to drive the trajectory towards the nominal waypoints. 
The acceleration term is fed forward in the control loop, and the position and velocity terms are
used as feedback to correct for errors, which is generally understood to provide better
tracking than PD control on its own~\cite{An1987-315}.
 The gravitational and Coriolis forces are included to counteract these effects from \autoref{eq:robot_dynamics}.
Finally we amplify the small scale $f$ with~$\Lambda=\text{diag}(450,450,45)$~to promote motion and map it to joint torques with
$J^{\top}\Lambda f$. The
purpose of this component is to shift the trajectory away from contact forces, with the goal of
correcting for misalignment. The gain values for $K_p$, $K_d$, and $\Lambda$ were chosen with
trial and error and could be refined in the future.

\subsection{Termination observer}
\label{subsec:term_obs}
The last component of the system is an observer to estimate when the swab has reached the
nasopharynx, during which the insertion stage should be terminated. Typically this would be followed
with rotating the swab to collect samples, but this motion is deferred for this paper. With the
absence of additional sensors, the controller makes this determination based on two sources of
information a) the Z-axis force and b) the total positional displacement of the
end-effector. Fuzzy logic \cite{Zadeh1978-353} presents a robust way of making this decision, which
 has seen use in situations where decisions rely on measurements that are linked to a state
 with uncertainty~\cite{Silva1995-351}~\cite{Saffiotti1997-352}. The fuzzy model we use is
\begin{equation}
  \begin{aligned}
  p_f &= \sigma (\nu_f(f_z - \bar{f}_z)))\\
  p_\epsilon &= \sigma (\nu_\epsilon(\epsilon - \bar{\epsilon}))\\
  p_{\text{term}} &= p_f p_\epsilon ,
  \label{eq:term}
  \end{aligned}
\end{equation}
where $\sigma$ is the sigmoid function and the termination decision $p_{\text{term}}$ is activated when both the terms for force, $p_f$,
and position, $p_\epsilon$, are sufficiently saturated. The force sigmoid $p_f$ saturates when $f_z$
rises, which we expect to happen when the swab tip makes contact with the nasopharynx. Likewise, the
position sigmoid $p_\epsilon$ saturates when the total displacement of the end-effector position
from its starting position, $\epsilon$, is high enough so that contact with the nasopharynx is
possible. The parameters $\nu_f$, $\nu_\epsilon$, $\bar{f}_z$, and $\bar{\epsilon}$ control the
scale and intercept of the sigmoid activations. We set the threshold for $p_{\text{term}}$ at 0.5,
where if at any point $p_{\text{term}}>0.5$, the trajectory halts. We set $\bar{f}_z =0.167$, $\nu_f =
30$, $\bar{\epsilon} = 0.085$ m, and $\nu_\epsilon = 40$. Intercept $\bar{\epsilon}$ was chosen
based on the nasal cavity geometry as we expect the swab to travel at least 9 cm. The other
parameters were chosen by trial and error; these values could also be refined because we typically
saw $p_f$ become saturated before $p_\epsilon$.
\begin{table}\vspace{0.2cm}
  \caption{Contingency table comparing the rates of success between the two controllers. A chi-square test results in a p-value = $2.49 \times 10^{-4}$, indicating that the proposed method has a significantly higher success rate.}
  \label{tab:success}
 \begin{center}
   \begin{tabular}{|c|c|c|}\hline
    Controller &  Success & Failure \\\hline
    Force-feedback       & 33 & 9 \\
    Baseline      & 16 & 25\\\hline
  \end{tabular}
  \end{center}
\end{table}

\begin{table}
  \caption{Summary of measured forces applied during the swab over all trials. Peak force and the forces averaged over the duration of the insertion are compared between the two controllers using paired t-tests.}
  \label{tab:force}
  \begin{center}
\begin{tabular}{|c|c|c|c|}\hline
 &Force-feedback(mN)&Baseline(mN)&p-value\\\hline
Avg. Force & $250.1 \pm 111.5$ & $374.7 \pm 209.8$ & $4.6 \times 10^{-5}$ \\\hline
Peak Force & $1063.8 \pm 324.5$ & $1347.3 \pm 515.9$ & $5.3 \times 10^{-4}$ \\\hline
\end{tabular}
\end{center}
\end{table}

\section{Experiments}
\label{sec:experiments}
The goal of the following experiments is to evaluate how well the proposed force-feedback controller can insert the
swab into the nasal cavity phantom described in \autoref{subsec:apparatus}. We consider the scenario
where the arm is already positioned and oriented in front of the nose and is ready to be
inserted. We also ignore the latter stages of the test that would typically involve swab rotation,
extraction, and sample preparation. Two different controllers are examined: the proposed controller with force-feedback and a
baseline controller that does not use force feedback. Specifically, the baseline controller uses the
same control law (\autoref{eq:torque_control}) but sets $\Lambda=0$, ignores force for determining
waypoints (i.e., $\dot{s}=1$), but still uses the same termination condition
\autoref{eq:term}. Comparisons are made between the two controllers to test whether our proposed
control method improves upon solely following the nominal trajectory.

\begin{figure}
  \includegraphics[width=\linewidth]{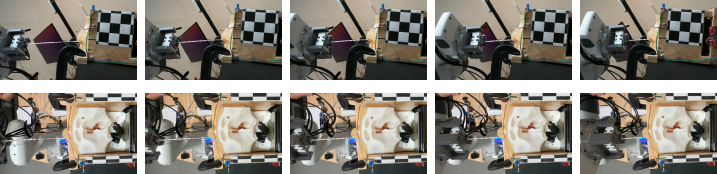}
  \caption{Side and top frames of a successful insertion taken by the two GoPro cameras.}
  \label{fig:frames}
\end{figure}

\begin{figure}
  \begin{minipage}{0.48\linewidth}
\centering \includegraphics[width=\linewidth]{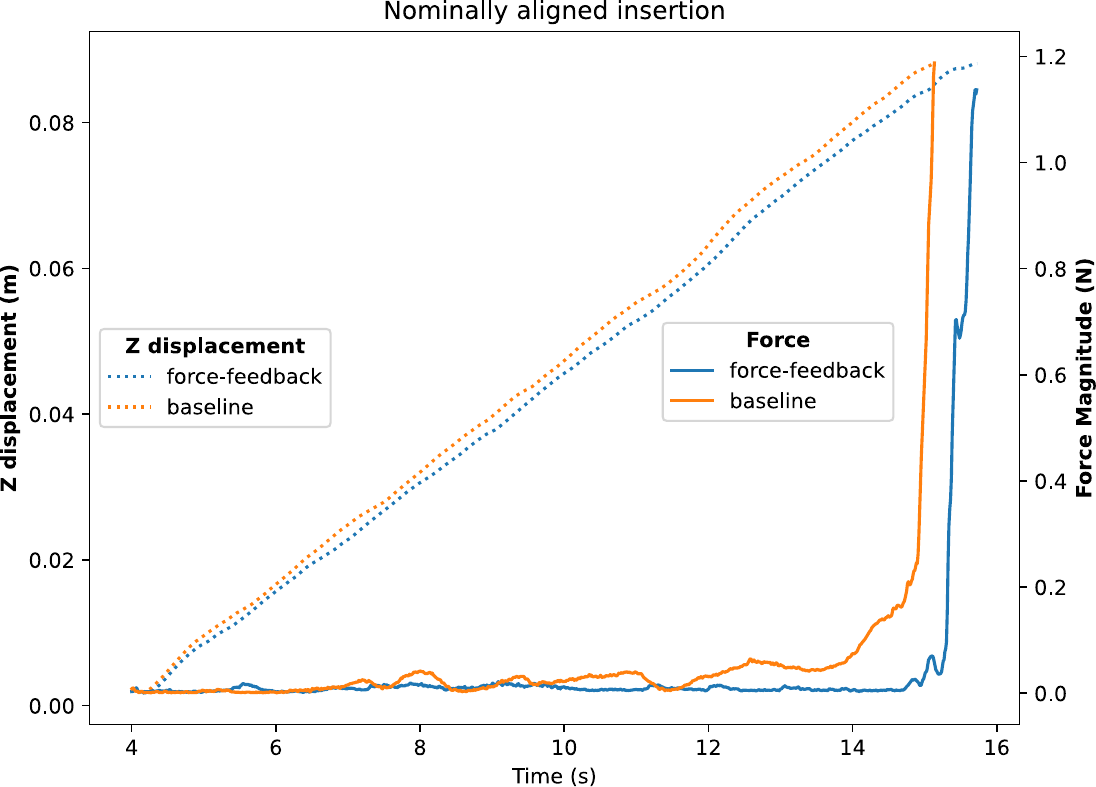} \\ (a)
    \end{minipage} \hfill \begin{minipage}{0.48\linewidth}\centering\includegraphics[width=\linewidth]{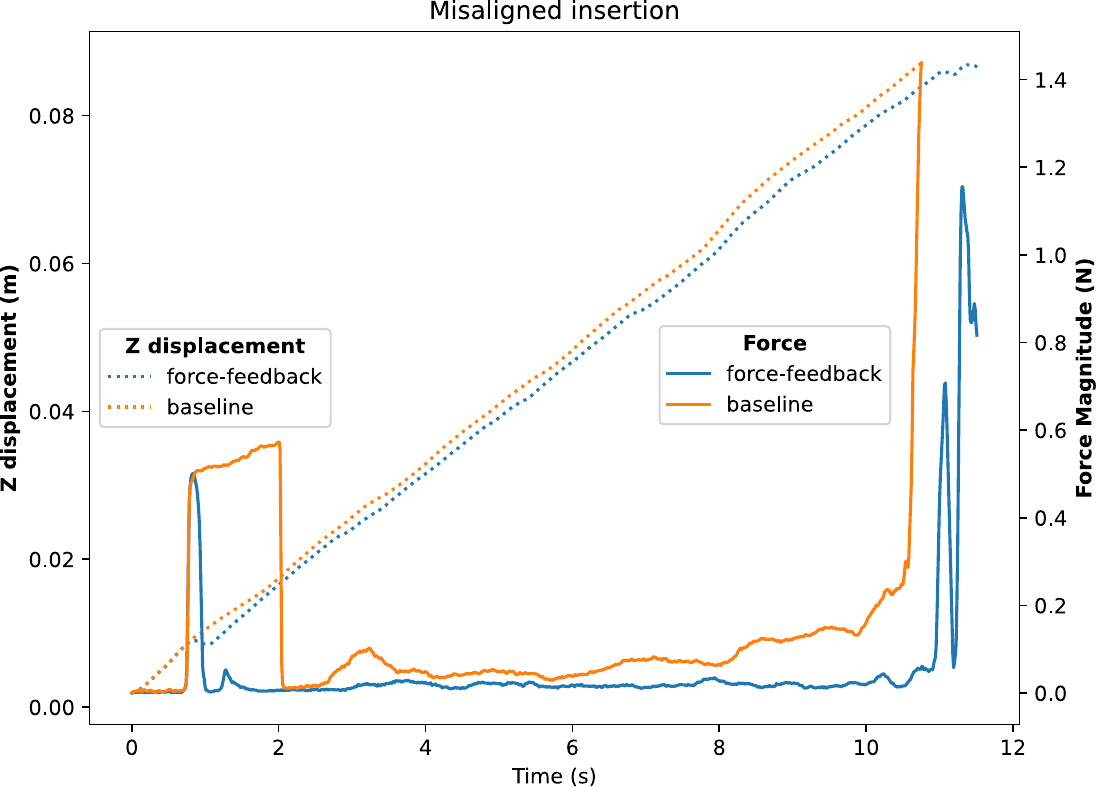}\\ (b)\end{minipage}
  \caption{Comparison of total observed forces and the accompanying displacement along the
      insertion axis for both versions of controllers recorded from the start of motion until
      the termination condition is reached, which is marked by the impulse of force as the swab reaches the nasopharynx. a) forces
    encountered during a nominal insertion angle. b) forces encountered during a misaligned
    insertion. Notice how in the bottom graph the proposed controller is able to adjust and minimize
  a collision that occurs early in the insertion.}
  \label{fig:force_graph}
\end{figure}

The insertion trials were done by moving the arm to the first waypoint, after which the apparatus was
manually positioned in front of the swab. The controllers were evaluated in a paired manner, where
the force-feedback controller would execute, then after resetting the arm to the starting position
without moving the apparatus, the baseline controller would execute. We executed a total of 41 of
these paired trials. Between each of these trials we would move the apparatus to a unique position and
angle in order to add variance and also to simulate pose estimation errors that would occur if a visual
system were guiding the initial placement. Fig. \ref{fig:frames} shows frames from top and side videos that were taken during a
successful insertion. Typically the insertion would take about 15 seconds, but the total time varied
depending on the disturbances encountered from the misalignment. The video attached in the
supplementary data includes a couple of recordings taken during the insertion trials.

Data recorded in the trials consists of videos from the two cameras and the published robot states,
loadcell, and FSR values in the ROS server. Success was defined based on if the swab reached
the nasopharynx and was largely determined by visual examination of the swab because the threshold of
force needed activate the FSR was too high. As shown in \autoref{tab:success}, the trials resulted
in a higher insertion success rate of 78.6\% for the force-feedback controller compared to the
baseline controller with 39.0\%. Comparing these outcomes with a chi-square test shows that the result
is statistically significant $p=2.49 \times 10^{-4} < 0.05$. It is also interesting to examine the
amounts of force applied to the swab, as this can be used as a proxy to the quality of the insertion
and to the patient's comfort. The forces that were sustained on the loadcell during trajectories are
shown in \autoref{tab:force}. In a paired manner, the force-feedback controller sustained less
forces than the baseline controller, but there was wide inter-trial variance based on the swab's
initial
placement. Fig. \ref{fig:force_graph} shows two graphs comparing the forces and displacement observed by the
controllers on two different initial alignments, which run until the termination observer triggers from sufficient force and displacement. The left graph shows the forces for a nominally
positioned swab, and the right graph shows the forces for a misaligned swab. In the latter case, the
force-feedback controller is able to adjust itself and sustain less persistent force compared to the baseline. From examining the final end-effector pose from forward kinematics of the recorded joint angles, the impact of the force feedback is also apparent: the trajectory from the left graph had a displacement of 2.4 mm and 0.77 degrees, while the right graph was altered by a larger factor with a displacement of 6.9 mm and 2.1 degrees.

\section{Discussion}
\label{sec:discussion}

As we showed in
\autoref{tab:success}, the baseline controller was significantly more successful than the force-feedback
controller according to a Chi-squared test, suggesting that incorporating force in the control loop enabled the insertion
trajectories to be successfully altered to non-ideal alignments with respect to the phantom. The
lower sustained forces in the force-feedback controller also indicates that it performed better than
the baseline. The peak force was about 0.4 N higher than the expert practitioner's from
Park \textit{et al.}~\cite{Park2021-301}, which could stem from physical differences in the phantom or from them only using
nominal insertion angles, but could also indicate that our controller parameters in \autoref{eq:s}
and \autoref{eq:term} could be better tuned in the future. Hence, the outcome of these trials
  show that there is certainly benefit to incorporating such a force-sensed based compliant control
  system for the proposed robotic arm setup that will increase insertion success and patient comfort versus a
closed loop position controller.

  Generally the compliance of the swab meant that there was significant allowance for the swab to be
  misaligned and still make it to the nasopharynx. Qualitatively, we notice that the pose and ease of insertion seemed to correlate with
  the findings from previous work \cite{Lee2022-309}. Insertion angles that were oriented towards the septum were
  generally more successful than those oriented away from the septum. Being positioned away from the nasal
  vestibule wall also seemed to be important for avoiding the wedging state and to reduce strain on
  the swab. While characteristics may differ for individual anatomy, we plan to take these
  observations into account when we design the vision guided positioning system for the initial
  placement of the swab in future work.

\begin{figure}[!htb]
  \includegraphics[width=\linewidth]{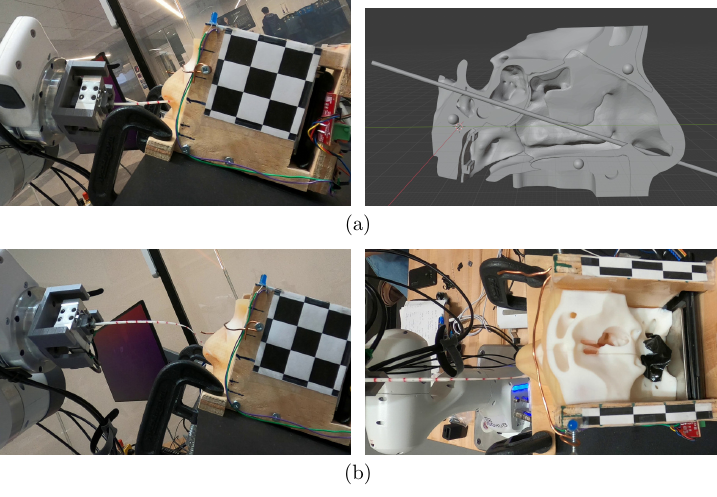}
  \caption{Two types of failure states observed in the trials: a) The elevation angle for the
    insertion is too high, resulting in it travelling through the wrong passage in the nasal
    cavity. The proposed compliant controller prevented these states.  b) The swab becomes
    wedged before entering the nasal cavity. This state occurred with both controllers.}
        \label{fig:failure_ex}
      \end{figure}

It is insightful to examine the cases where the insertion failed for either controller. There were
two types of failure states that occurred during insertion.  The first type of
failure was unique to the baseline controller, which failed to reach the nasopharynx because the elevation angle of the swab
was too high (see Fig. \ref{fig:failure_ex} a). Typically this transpired when the swab was
positioned too low and then became levered into an excessive angle where it deflected off the
sphenoid sinus, which would likely be more uncomfortable and have higher chance of complications for
a human patient. The force-feedback controller was able to adjust the trajectory to avoid this
levering effect because of the adjustments the force-feedback produced.
The swab becoming stuck and being unable to enter the nasal cavity, as shown in
Fig. \ref{fig:failure_ex} b), was the second type of failure. This
failure state was the result of the swab becoming wedged on the nasal vestibule because of poor
alignment and appeared with both the force-feedback and baseline controller. This highlights
  that implementing a strategy to detect contact within the nasal vestibule and to compensate the
  trajectory of the swab during the first centimetre of insertion as the main
  recommendation to improve insertion success.

  In terms of future work, one of the major aspects that would need to be resolved is enforcing
  guarantees for safety. One such area is providing guarantees for the stability of the
  controller. While the force filter in \autoref{subsec:force_filter} helped stabilize the response
  compared to using unfiltered values, there were still some cases where oscillations were present
  as the swab reached the nasopharynx. An untested scenario is the controller's response to the
  natural motion that a human patient would have during the procedure. From the hardware design, it
  is important to eventually build in a mechanism to detach the swab from the robot in cases where
  extremely high forces are detected or when the participant wants to abort the procedure. Finally,
  although the force-feedback controller was able to handle non-ideal insertion poses, it may be
  fruitful to investigate if applying an estimator that directly estimates pose errors via the force
  feedback could result in improvement, particularly for resolving the wedging failure case.

\section{Conclusion}
\label{sec:conclusion}
In this work, we proposed a scenario where a standardized fixed rigid arm robot performs NP swab
sampling using a compliant control system. To investigate this, we designed a minimal force sensing
end-effector and integrated it into a Panda arm that could be adapted to similar rigid arms. We
designed an admittance force-feedback torque control system to perform the insertion test. We
performed experiments to evaluate our system on a 3D printed nasal cavity from variety of different
alignment conditions and showed that the admittance controller was succeeded at a rate of 78.6\%
compared to 39.0\% of the baseline position controller. This demonstrates that there is feasibility
for a rigid arm to perform the NP swab test on people using sensitive force feedback as a
modality. The compliant controller was able to compensate for some misalignment and
thereby avoid some failure states, however it is clear
that a full solution will require additional sensors or other strategies to adjust for more extreme
misalignment and more study will be needed to evaluate the impact of head motion on the controller.

Future work will extend this research towards implementing a fully automated
  robotic NP sampling solution. An eye-in-hand visual servo system will be a necessary for reaching
  the pose in front of the nostril, prior to insertion. The other stages of the NP test
  (rotating the swab at the nasopharynx and extraction) will need to be implemented. As well, adding
  additional inputs, such as tracked visual features of the face, may be worth fusing 
  with the force measurements as inputs to the control system to better
  adjust to disturbances such as motions of the head and general misalignment.

While the NP swab test is just one task a healthcare worker would do in a clinical setting, there
are many other types of routine close-contact tasks that would require similar levels of dexterity and force
sensitivity. Examples include other types of sample collection, medication or blood collection
through needles, and skin temperature testing. Consequently, having a multi-purpose robot that could do NP swab tests could open up a
multitude of different clinical tasks, and could be a boon for the healthcare system by allowing
procedures to be done autonomously, enabling human resources to be reallocated.

\section*{ACKNOWLEDGMENT}
This research received funding from the Natural Sciences and Engineering Research Council of Canada
(NSERC), from the Tri-Agency Canada Excellence Research Chair Program, and from the University of
Waterloo. Thanks are extended to the University of Waterloo Robohub for use of their Panda arm.

\section*{Author contributions}
P.L. designed materials, conducted experiments, drafted manuscript. J.Z. and K.M. supervised work and
drafted manuscript.

\section*{Additional Information}
\textbf{Competing Interests:} The authors report no competing interests.

\section*{Data availability}
Materials used in this study is included in the supplementary files.

\printbibliography

\end{document}